\pdfoutput=1

\documentclass[11pt]{article}

\usepackage{acl}

\usepackage{times}
\usepackage{latexsym}

\usepackage[T1]{fontenc}

\usepackage[utf8]{inputenc}

\usepackage{microtype}

\usepackage[mode=buildnew]{standalone}
\usepackage{algorithm}
\usepackage{algorithmic}
\usepackage{booktabs}
\usepackage{subcaption}
\usepackage{arydshln}
\usepackage{enumitem}

\usepackage{todonotes}

\usepackage{amsmath,amsfonts,bm}

\def\eqref#1{equation~\ref{#1}}

\def\1{\bm{1}}

\def\vc{{\bm{c}}}

\def\vq{{\bm{q}}}

\def\vt{{\bm{t}}}

\def\mI{{\bm{I}}}

\DeclareMathAlphabet{\mathsfit}{\encodingdefault}{\sfdefault}{m}{sl}
\SetMathAlphabet{\mathsfit}{bold}{\encodingdefault}{\sfdefault}{bx}{n}

\newcommand{\R}{\mathbb{R}}

\newcommand{\sigmoid}{\sigma}

\newcommand{\question}{Q}

\newcommand{\sql}{P}

\newcommand{\schema}{\mathcal{S}}
\renewcommand{\ast}{T}

\newcommand{\sDatabase}{\mathcal{D}}
\newcommand{\sSourceDomain}{\mathcal{D}_s}
\newcommand{\sTargetDomain}{\mathcal{D}_t}
\newcommand{\sSourceDomainBatch}{\mathcal{B}_s}
\newcommand{\sTargetDomainBatch}{\mathcal{B}_t}
\newcommand{\sBatch}{\mathcal{B}}

\newcommand{\param}{\boldsymbol \theta}
\newcommand{\update}{\text{Update}}
\newcommand{\attention}{\text{Attention}}
\newcommand{\mlp}{\text{MLP}}
\newcommand{\loss}{\mathcal{L}}

\newcommand{\comment}[1]{}

\newcommand\enspidermargin{2.1\%}
\newcommand\chspidermargin{4.5\%}

\title{Meta-Learning for Domain Generalization in Semantic Parsing}

\author{Bailin Wang, Mirella Lapata \and Ivan Titov \\
  Institute for Language, Cognition and Computation \\ 
  School of Informatics, University of Edinburgh \\
  {\tt bailin.wang@ed.ac.uk, \tt \{mlap, ititov\}@inf.ed.ac.uk}}

\date{}

\begin{document}

\maketitle
\begin{abstract}
The importance of building semantic parsers which can be 
applied to new domains and generate programs unseen at training has long been acknowledged, 
and datasets  testing out-of-domain performance are becoming increasingly available.
However, little or no attention has been devoted to learning algorithms or objectives which promote domain generalization, 
with virtually all existing approaches relying on standard supervised learning.
In this work, we use a meta-learning framework 
which targets zero-shot domain generalization for semantic parsing.
We apply a model-agnostic training algorithm that simulates zero-shot parsing by 
constructing virtual train and test sets from disjoint domains. 
The learning objective capitalizes on the intuition that gradient steps that improve source-domain performance 
should also improve target-domain performance, 
thus encouraging a parser to generalize to unseen target domains.
Experimental results on the (English) Spider and Chinese Spider datasets show that 
the meta-learning objective significantly boosts the performance of a baseline parser. 
\end{abstract}

\section{Introduction}

Semantic parsing is the task of mapping natural language (NL) utterances to executable programs.  
While there has been much progress in this area, earlier work has primarily focused on evaluating parsers in-domain (e.g., tables or databases) 
and often with the same programs as those provided in training~\cite{finegan-dollak-etal-2018-improving}. 
A much more challenging goal is achieving {\it domain generalization}, i.e.,~building parsers
which can be successfully applied to new domains and  are able to produce complex unseen programs. 
Achieving this generalization goal would, in principle, let users
query arbitrary (semi-)structured data on the Web and reduce the
annotation effort required to build multi-domain NL interfaces
(e.g.,~Apple Siri or Amazon Alexa).  Current parsers struggle in this
setting; for example, we show in Section~\ref{sec:experiments} that
a modern parser trained on the challenging Spider
dataset~\cite{yu-etal-2018-spider} has a gap of more than 25\% in
accuracy between in- and out-of-domain performance.  While the
importance of domain generalization has been previously
acknowledged~\cite{cai-yates-2013-large,chang2019zeroshot}, and
datasets targetting  {\it zero-shot} (or out-of-domain) performance
are becoming increasingly
available~\cite{pasupat-liang-2015-compositional,wang-etal-2015-building,zhongSeq2SQL2017,yu-etal-2018-spider},
little or no attention has been devoted to studying learning
algorithms or objectives which promote domain generalization.

\begin{figure}[t]
\centering
\includegraphics[width=0.46 \textwidth]{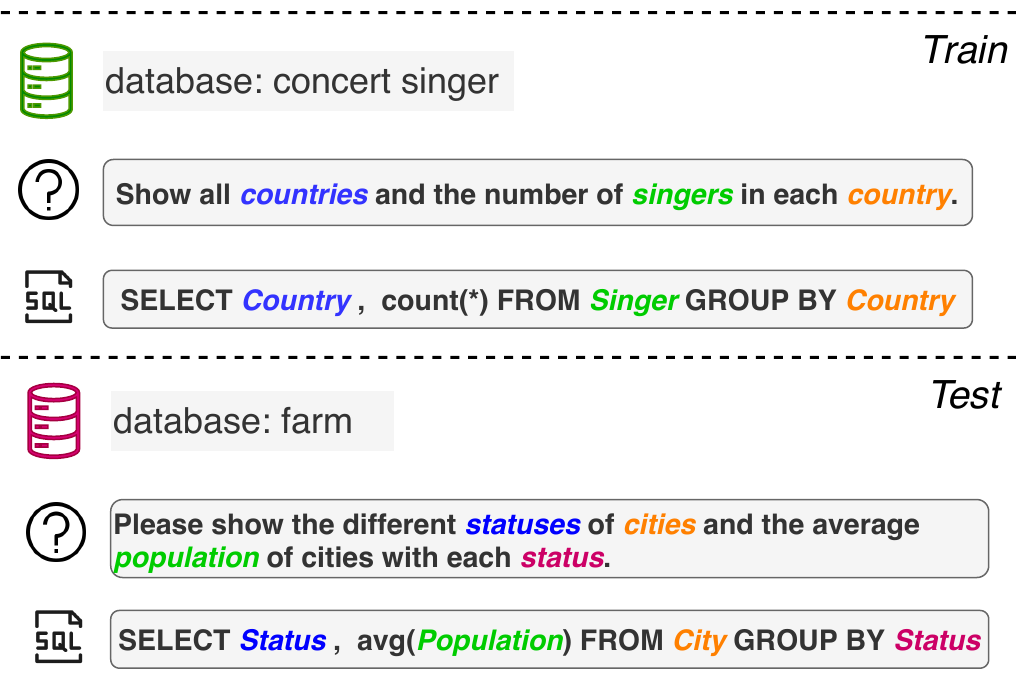}
\caption{Zero-shot semantic parsing: at training time, a parser observes instances for the database \textit{concert singer}. 
At test time, it needs to generate SQL for questions pertaining to the unseen database \textit{farm}.}
\label{fig:example}
\vspace{-5mm}
\end{figure}

\comment{
Semantic parsing is the task of mapping natural language (NL) utterance to meaning representation that are executable in various domains.
It serves as a powerful paradigm for ubiquitous NL interfaces, e.g., Apple Siri, Amazon Alexa.
In real-life applications, these NL interfaces need to handle queries in a large amount of domains.
Hence, scaling up a semantic parser to unseen and complex domains has received increasing interest in recent years.
Recent works focus on making semantic parsers that can access a wide range of (semi-)structured data, 
like web tables~\cite{pasupat-liang-2015-compositional}, knowledge bases~\cite{wang-etal-2015-building} or relational databases~\cite{zhongSeq2SQL2017}.
The challenging setting of zero-shot semantic parsing,
where a parser needs to generalize to unseen domains at test time, 
has emerged to examine the \textit{domain generalization} of a parser.
A representative task we focus on is zero-shot text-to-SQL parsing 
where we aim at translating NL questions to SQL queries and conduct evaluations in unseen databases. 
Consider the example in Figure \ref{fig:example}, a parser needs to process questions from a new database at test time.}

Conventional supervised learning simply assumes that source- and
target-domain data originate from the same distribution, and as a
result struggles to capture this notion of domain generalization for
zero-shot semantic parsing.  Previous approaches
\cite{guo-etal-2019-towards,wang2019rat,herzig-berant-2018-decoupling}
facilitate domain generalization by incorporating inductive biases in
the model, e.g., designing linking features or functions which should
be invariant under domain shifts.  In this work, we take a different
direction and improve the domain generalization of a semantic parser by
modifying the learning algorithm and the objective.  We draw
inspiration from  meta-learning 
\cite{finn2017model,li2018maml} and use an objective that optimizes
for domain generalization.
That is, we 
consider a set of tasks, where each task is a zero-shot semantic parsing task with its own source and target domains. 
By optimizing towards better target-domain performance on each task, 
we encourage a parser to extrapolate from source-domain data and achieve better domain generalization. 

Specifically, we focus on text-to-SQL parsing where we aim at
translating NL questions to SQL queries and conduct evaluations on
unseen databases.  Consider the example in Figure \ref{fig:example}, a
parser needs to process questions to a new database at test time.  To
simulate this scenario during training, we synthesize a set of virtual
zero-shot parsing tasks by sampling disjoint source and target domains\footnote{We use the terms domain and database interchangeably.}
for each task from the training domains.  The objective we require is
that gradient steps computed towards better source-domain performance
would also be beneficial to target-domain performance.  One can think
of the objective as consisting of both the loss on the source domain
(as in standard supervised learning) and a regularizer, equal to the
dot product between gradients computed on source- and target-domain
data.  Maximizing this regularizer favours finding model parameters
that work not only on the source domain but also generalize to
target-domain data.  The objective is borrowed from \citet{li2018maml}
who adapt a Model-Agnostic Meta-Learning (MAML;
\citealt{finn2017model}) technique for domain generalization in
computer vision.  In this work, we study the effectiveness of this
objective in the context of semantic parsing. 
This objective is model-agnostic, simple to incorporate and does not require any changes in the parsing model itself. 
Moreover, it does not introduce new parameters for meta-learning.

Our contributions can be summarized as follows.
%
\begin{itemize}
\item We handle zero-shot semantic parsing by applying a meta-learning
  objective that \emph{directly} optimizes for domain generalization.

    \item We propose an approximation of the meta-learning objective 
    that is more efficient and allows more scalable training. 

    \item We perform experiments on two text-to-SQL benchmarks: Spider and Chinese Spider.
      Our new training objectives obtain significant improvements in accuracy
      over a baseline parser trained with conventional supervised
      learning.


    \item We show that even when parsers are augmented with pre-trained models, e.g., BERT,
      our method can still effectively improve domain generalization in terms of accuracy.

\end{itemize}
%
    Our code is available at \url{https://github.com/berlino/tensor2struct-public}.




\section{Related Work}
\paragraph{Zero-Shot Semantic Parsing}
\label{ssec:zsp_compare}

Developing a parser that can generalize to unseen domains has attracted
increased attention in recent years.  
Previous work has mainly focused on the sub-task of \textit{schema
linking} as a means of promoting domain generalization.  In schema
linking, we need to recognize which columns or tables are mentioned in
a question.  For example, a parser would decide to select the column
\textit{Status} because of the word \textit{statuses} in
Figure~\ref{fig:example}.  However, in the setting of zero-shot
parsing, columns or tables might be mentioned in a question without
ever being observed during training.  

One line of work tries to incorporate
inductive biases, e.g., domain-invariant n-gram matching features
\cite{guo-etal-2019-towards,wang2019rat}, cross-domain alignment
functions \cite{herzig-berant-2018-decoupling}, or auxiliary linking
tasks~\cite{chang2019zeroshot} to improve schema linking.  However, in
the cross-lingual setting of Chinese Spider
\cite{min-etal-2019-pilot}, where questions and schemas are not in the
same language, it is not obvious how to design such inductive biases like
n-gram matching features.  Another line
of work relies on large-scale unsupervised pre-training on massive
tables \cite{herzig2020tapas,yin20acl} to obtain better
representations for both questions and database schemas.  Our work is
orthogonal to these approaches and can be easily coupled with them.
As an example, we show in Section \ref{sec:experiments} that our
training procedure can improve the performance of a parser already
enhanced with n-gram matching features
\cite{guo-etal-2019-towards,wang2019rat}. 

Our work is similar in spirit to \citet{givoli-reichart-2019-zero}, 
who also attempts to simulate source and target domains during learning.
However, their optimization updates on virtual source and target domains 
are loosely connected by a two-step training procedure where a parser is 
first pre-trained on virtual source domains and then fine-tuned on virtual 
target domains. As we will show in Section~\ref{sec:maml}, our training procedure 
does not fine-tune on virtual target domains but rather, uses them to evaluate a 
gradient step (for every batch) on source domains. This is better aligned with what is expected of the parser at test time: 
there will be no fine-tuning on  \emph{real} target domains at test time
so there should not be any fine-tuning on \emph{simulated} ones at train time either. 
Moreover, \citet{givoli-reichart-2019-zero} treat the division of training domains to 
virtual train and test domains as a hyper-parameter, 
which is possible for a handful of domains, but problematic when dealing
with hundreds of domains as is the case for text-to-SQL parsing.

\paragraph{Meta-Learning for NLP}
Meta-learning has been receiving soaring interest in the machine learning
community.  Unlike conventional supervised learning, meta-learning
operates on tasks, instead of data points.  Most previous work
\cite{vinyals2016matching,ravi2016optimization, finn2017model} has
focused on few-shot learning where meta-learning helps address the
problem of learning to learn fast for adaptation to a new task or
domain.  Applications of meta-learning in NLP are cast in a similar vein and include
machine translation
\cite{gu-etal-2018-meta} and relation classification
\cite{obamuyide-vlachos-2019-model}.    
The meta-learning framework however is more general,  with the algorithms or underlying ideas applied, e.g.,~to continual learning~\cite{gupta2020maml}, semi-supervised learning~\cite{ren2018meta}, multi-task learning
~\cite{yu2020gradient} and, as in our case, domain generalization~\cite{li2018maml}. 

Very recently, there have been some applications of MAML to semantic
parsing tasks \cite{huang-etal-2018-natural,guo-etal-2019-coupling,
sun2019neural}.  These approaches simulate {\it few-shot} learning
scenarios in training by constructing a pseudo-task for each example. 
Given an example, similar examples are retrieved from the original training set.
MAML then encourages strong performance on the retrieved examples after an update on the original example, simulating test-time fine-tuning.
 \citet{lee2019oneshot} use matching
networks~\cite{vinyals2016matching} to enable {\it one-shot}
text-to-SQL parsing where tasks for meta-learning are defined by SQL
templates, i.e., a parser is expected to generalize to a new SQL
template with one example.  In contrast, the tasks we construct for
meta-learning aim to encourage {\it generalization} across domains,
instead of {\it adaptation} to a new task with one (or few)
examples. One clear difference lies in how meta-train and meta-test sets
are constructed. In previous work (e.g.,
\citealt{huang-etal-2018-natural}), these come from the \emph{same}
domain whereas we simulate domain \emph{shift} and sample different
sets of domains for meta-train and meta-test.

\paragraph{Domain Generalization}
Although the notion of domain generalization has been less explored in
semantic parsing, it has been 
studied in other areas such as 
computer vision \cite{ghifary2015domain,zaheer2017deep,li2018domain}.
Recent work \cite{li2018maml,metareg2018} employed optimization-based
meta-learning to handle domain shift issues in domain generalization.
We employ the meta-learning objective originally proposed in
\citet{li2018maml}, where they adapt MAML to encourage generalization
in unseen domains (of images).  Based on this objective, we propose a
cheap alternative that only requires first-order gradients, thus
alleviating the overhead of computing second-order derivatives required
by MAML.

\section{Meta-Learning for Domain Generalization}
\label{sec:maml}
We first formally define the problem of domain generalization
in the context of zero-shot text-to-SQL parsing.
Then, we introduce DG-MAML, a training algorithm that 
helps a parser achieve better domain generalization.
Finally, we propose a computationally cheap approximation thereof.

\subsection{Problem Definition}
\paragraph{Domain Generalization} 
Given a natural language question $\question$
in the context of a relational database $\sDatabase$,
we aim at generating the corresponding SQL $\sql$.
In the setting of zero-shot parsing, we have a set of source domains
$\sSourceDomain$ where labeled question-SQL pairs are available. 
We aim at developing a parser that can perform well on a set of unseen target domains $\sTargetDomain$.
We refer to this problem as \textit{domain generalization}.

\paragraph{Parsing Model}
We assume a parameterized parsing model that specifies a predictive distribution 
$p_{\param}(\sql|\question, \sDatabase)$ over all possible SQLs.
For domain generalization, a parsing model needs to properly condition on its input of questions 
and databases such that it can generalize well to unseen domains.

\paragraph{Conventional Supervised Learning} 
Assuming that question-SQL pairs from source domains and target domains are sampled 
i.i.d from the same distribution, the typical training objective of 
supervised learning is to minimize the loss function of the negative log-likelihood of the gold SQL query:
\begin{equation}
    \loss_{\sBatch}(\param) = - \frac{1}{N} \sum_{i=1}^{N}  \log p_{\param}(\sql| \question, \sDatabase) 
\end{equation}
where $N$ is the size of mini-batch $\sBatch$.
Since a mini-batch is randomly sampled from all training source domains $\sSourceDomain$, 
it usually contains question-SQL pairs from a mixture of different domains.

\paragraph{Distribution of Tasks}
Instead of treating semantic parsing as a conventional supervised
learning problem, we take an alternative view based on
meta-learning.  Basically, wea re interested in a learning algorithm that can benefit
from a distribution of choices of source and target domains, denoted by $p(\tau)$, where $\tau$ refers to an
instance of a zero-shot semantic parsing task that has its own source
and target domains. 

In practice, we usually have a fixed set of training source domains
$\sSourceDomain$.  We construct a set of virtual tasks $\tau$
by randomly sampling disjoint source and target domains from the
training domains.  Intuitively, we assume that divergences between
the test and training domains during the learning phase are representative of
differences between training and actual test domains.  This is
still an assumption, but considerably weaker compared to the
i.i.d. assumption used in conventional supervised learning.
Next, we introduce the training algorithm called DG-MAML motivated by this
assumption.

\subsection{Learning to Generalize with DG-MAML}

Having simulated source and target domains for each virtual task, we
now need a training algorithm that encourages generalization to unseen
target domains in each task.  For this, we turn to optimization-based
meta-learning algorithms \cite{finn2017model,nichol2018first,li2018maml} 
and apply DG-MAML (Domain Generalization with Model-Agnostic Meta-Learning), 
a variant of MAML \cite{finn2017model} for this purpose.  Intuitively, DG-MAML
encourages the optimization in the source domain to have a positive
effect on the target domain as well. 

During each learning episode of DG-MAML, we randomly sample a task $\tau$
which has its own source domain $\sSourceDomain^\tau$ and target domain $\sTargetDomain^\tau$.
For the sake of efficiency, we randomly sample  mini-batch question-SQL pairs 
$\sSourceDomainBatch$ and $\sTargetDomainBatch$  from $\sSourceDomain^\tau$ and $\sTargetDomain^\tau$, respectively, 
for learning in each task.
DG-MAML conducts optimization in two steps, namely \textit{meta-train}
and \textit{meta-test}.  

\paragraph{Meta-Train} 
DG-MAML first optimizes parameters towards better performance in the virtual source domain $\sSourceDomain^\tau$
by taking one step of stochastic gradient descent (SGD) from the loss under $\sSourceDomainBatch$.
\begin{equation}
    \param'\leftarrow \param - \alpha \nabla_{\param} \loss_{\sSourceDomainBatch}  (\param) 
    \label{eq:meta_train}
\end{equation}
where $\alpha$ is a scalar denoting the learning rate of meta-train.
This step resembles conventional supervised learning where we use stochastic gradient descent to optimize the parameters. 

\paragraph{Meta-Test}

We then evaluate the resulting parameter $\param'$ in the virtual target domain $\sTargetDomain$ 
by computing the loss under $\sTargetDomainBatch$, 
which is denoted as $\loss_{\sTargetDomainBatch}(\param')$.

Our final objective for a task $\tau$ is to minimize the joint loss on $\sSourceDomain$ and $\sTargetDomain$:
\begin{align}
    \begin{split}
    \loss_{\tau}(\param) &= \loss_{\sSourceDomainBatch}  (\param) + \loss_{\sTargetDomainBatch}(\param') \\
       &= \loss_{\sSourceDomainBatch}  (\param) + \loss_{\sTargetDomainBatch}(
        \param - \alpha \nabla_{\param} \loss_{\sSourceDomainBatch}  (\param)) 
    \end{split}
\label{eq:obj}
\end{align}
where we optimize towards the better source \emph{and} target domain performance simultaneously. 
Intuitively, the objective requires that the gradient step conducted in the source domains in Equation~(\ref{eq:meta_train})
be beneficial to the performance of the target domain as well.
In comparison, conventional supervised learning, whose objective would be equivalent to 
$\loss_{\sSourceDomainBatch}  (\param) + \loss_{\sTargetDomainBatch}(\param)$, 
does not pose any constraint on the gradient updates.
As we will elaborate shortly, DG-MAML can be viewed as a regularization of gradient updates 
in addition to the objective of conventional supervised learning. 

We summarize our DG-MAML training process in Algorithm~\ref{algo:maml}.
Basically, it requires two steps of gradient update (Step \ref{algo:line_meta_train} and Step \ref{algo:line_meta_update}). 
Note that $\param'$ is a function of $\param$ after the meta-train update.
Hence, optimizing $\loss_{\tau}(\param)$ with respect to $\param$ 
involves optimizing the gradient update in Equation~(\ref{eq:meta_train})  as well. 
That is, when we update the parameters $\param$ in the final update of Step \ref{algo:line_meta_update}, 
the gradients need to back-propagate though the meta-train updates in Step \ref{algo:line_meta_train}. 
The update function in Step \ref{algo:line_meta_update} could be based on any gradient descent algorithm.
In this work  we use Adam \cite{kingma2014adam}.

\paragraph{Comment} 
Note that DG-MAML is different from MAML~\cite{finn2017model} which is typically used in the context of few-shot learning.
In our case, it encourages domain generalization during training, and does not require an adaptation phase.

\begin{algorithm}[t]
\caption{DG-MAML Training Algorithm}
\label{algo:maml}
\begin{algorithmic}[1]
  \REQUIRE Training databases $\sDatabase$
  \REQUIRE Learning rate $\alpha$
  \FOR{step $\gets 1$ \TO $T$}
    \STATE Sample a task $\tau$ of ($\sSourceDomain^\tau, \sTargetDomain^\tau)$  from $\sDatabase$ \\
    \STATE Sample mini-batch $\sSourceDomainBatch^\tau$  from  $\sSourceDomain^\tau$  \\
    \STATE Sample mini-batch $\sTargetDomainBatch^\tau$ from  $\sTargetDomain^\tau$  \\
    \STATE Meta-train update: \label{algo:line_meta_train}  \\ 
    \quad $\param'\leftarrow \param - \alpha \nabla_{\param} \loss_{\sSourceDomainBatch^\tau} (\param)$ \\ 
    \STATE Compute meta-test objective: \\ \quad $ \loss_{\tau}(\param) = \loss_{\sSourceDomainBatch}  (\param) + \loss_{\sTargetDomainBatch}(\param')$ \\
    \STATE Final Update: \label{algo:line_meta_update} \\ 
    \quad  $\param \leftarrow  \update(\param, \nabla_{\param} \loss_{\tau}(\param))$
   \ENDFOR
\end{algorithmic}
\end{algorithm}

\subsection{Analysis of DG-MAML}
\label{ssec:grad_analyses}

To give an intuition of the objective in Equation~(\ref{eq:obj}), we
follow previous work \cite{nichol2018first,li2018maml} and use the first-order
Taylor series expansion to approximate it:
\begin{align}
    \begin{split}
     \loss_{\tau}(\param) =& \loss_{\sSourceDomainBatch}  (\param) + \loss_{\sTargetDomainBatch}(\param') \\
      =&  \loss_{\sSourceDomainBatch}  (\param) + \loss_{\sTargetDomainBatch}(\param - \alpha \nabla_{\param} \loss_{\sSourceDomainBatch}(\param)) \\
     \approx &
        \loss_{\sSourceDomainBatch}(\param)  + \loss_{\sTargetDomainBatch}(\param) - \\
        & \alpha  (\nabla_{\param} \loss_{\sSourceDomainBatch}  (\param) \cdot \nabla_{\param} \loss_{\sTargetDomainBatch}  (\param))
    \end{split}
    \label{eq:maml_detailed_analysis}
\end{align}
where in the last step we expand the function
$\loss_{\sSourceDomainBatch}$ at~$\param$.  The approximated objective
sheds light on what DG-MAML optimizes. In addition to minimizing the
losses from both source and target domains, which are
$\loss_{\sSourceDomainBatch}(\param) +
\loss_{\sTargetDomainBatch}(\param)$, DG-MAML further tries to
maximize~$ \nabla_{\param} \loss_{\sSourceDomainBatch} (\param) \cdot
\nabla_{\param} \loss_{\sTargetDomainBatch} (\param)$, the dot product
between the gradients of source and target domain.  That is, it
encourages gradients to generalize between source and target domain
within each task $\tau$.

\subsection{First-Order Approximation}
\label{ssec:dg_fmaml}

The final update in Step \ref{algo:line_meta_update} of Algorithm~\ref{algo:maml} 
requires second-order derivatives, which may be problematic, inefficient or
non-stable with certain classes of models~\cite{mensch2018differentiable}.
Hence, we propose an approximation that only requires computing first-order derivatives.

First, the gradient of the objective in Equation~(\ref{eq:obj}) can be computed as:

\begin{align}
\begin{split}
   \nabla_{\param} \loss_{\tau}(\param) = &
    \nabla_{\param} \param' \nabla_{\param'} \loss_{\sTargetDomainBatch}(\param') + 
    \nabla_{\param} \loss_{\sSourceDomainBatch}(\param)  \\
    = & \big (\mI - \alpha  \nabla_{\param}^2  \loss_{\sSourceDomainBatch}(\param) \big ) 
    \nabla_{\param'} \loss_{\sTargetDomainBatch}(\param') \\ 
     & +\nabla_{\param} \loss_{\sSourceDomainBatch}(\param) 
\end{split}
\end{align}

\noindent where $\mI$ is an identity matrix and
\mbox{$ \nabla_{\param}^2 \loss_{\sSourceDomainBatch}(\param)$} is the
Hessian of $\loss_{\sSourceDomainBatch}$ at $\param$.  
We consider the alternative of ignoring this second-order term and simply assume that
\mbox{$\nabla_{\param} \param' = \mI$}.  In this variant, we simply
combine gradients from source and target domains.  We show in the
Appendix that this objective can still be viewed as maximizing the dot
product of gradients from source and target domain.

The resulting first-order training objective, which we refer to as
DG-FMAML, is inspired by Reptile, a first-order meta-learning
algorithm~\cite{nichol2018first} for few-shot learning.  A two-step
Reptile would compute SGD on the same batch twice while
DG-FMAML computes SGD on two different batches,
$\sSourceDomainBatch$ and $\sTargetDomainBatch$, once.  To put it
differently, DG-FMAML tries to encourage \emph{cross-domain} generalization
while Reptile encourages \emph{in-domain} generalization.

\section{Semantic Parser}
\label{sec:parser}
In general, DG-MAML is model-agnostic and can be coupled with any
semantic parser to improve its domain generalization.  In this work,
we use a base parser that is based on RAT-SQL~\cite{wang2019rat},
which currently achieves state-of-the-art performance on Spider.\footnote{
We re-implemented RAT-SQL, and added a component for value prediction so that 
our base parsers can be evaluated by execution accuracy.}

Formally, RAT-SQL takes as input question~$\question$ and
schema~$\schema$ of its corresponding database.  Then it produces a
program which is represented as an \textit{abstract syntax tree}
$\ast$ in the context-free grammar of
SQL~\cite{yin-neubig-2018-tranx}.  RAT-SQL adopts the encoder-decoder
framework for text-to-SQL parsing.  It  has three components:
an initial encoder, a transformer-based encoder and an LSTM-based
decoder.  The initial encoder provides initial representations,
denoted as $\question_{init}$ and $\schema_{init}$ for the question
and the schema, respectively.  A relation-aware transformer (RAT)
module then takes the initial representations and further computes
context-aware representations $\question_{enc}$ and $\schema_{enc}$
for the question and the schema, respectively.  Finally, a decoder
generates a sequence of production rules that constitute the abstract
syntax tree $\ast$ based on $\question_{enc}$ and $\schema_{enc}$.
To obtain $\question_{init}$ and $\schema_{init}$, the initial encoder
could either be 1) LSTMs~\cite{hochreiter1997long} on top of
pre-trained word embeddings, like
GloVe~\cite{jeffreypennington2014glove}, or 2) pre-trained contextual
embeddings like BERT~\cite{devlin2018bert}.  In our work, we will test
the effectiveness of our method for both variants.

As shown in \citet{wang2019rat}, the encodings $\question_{enc}$
and $\schema_{enc}$, which are the output of the RAT module, 
heavily rely on schema-linking features.  These features are extracted from
a heuristic function that links question words to  columns and
tables based on n-gram matching, and they are readily available in
the conventional mono-lingual setting of the Spider dataset. 
However, we hypothesize that the  parser's over-reliance on these features 
is specific to Spider, where annotators were shown the database schema and
asked to formulate queries. As a result, they were prone to re-using terms from 
the schema verbatim in their questions. This would not be the case in a real-world application 
where users are unfamiliar with the structure of the underlying database and free to 
use arbitrary terms which would not necessarily match column or table names~\cite{suhr-etal-2020-exploring}. 
Hence, we will also evaluate our parser in the cross-lingual setting where $\question$ and $\schema$ are
not in the same language, and such features would not  be available.   

\section{Experiments}
\label{sec:experiments}
To evaluate DG-MAML, we integrate it with a base
parser and test it on zero-shot text-to-SQL tasks. By designing an in-domain benchmark, we
also show that the out-of-domain improvement does not come at the cost
of in-domain performance. We also present
some analysis to show how DG-MAML affects domain
generalization.

\subsection{Datasets and Metrics}
We evaluate DG-MAML on two zero-shot text-to-SQL benchmarks, namely,
(English) Spider~\cite{yu-etal-2018-spider} and  Chinese
Spider~\cite{min-etal-2019-pilot}.  
Chinese Spider is a Chinese version of Spider that
translates all  NL questions  from English to Chinese and
keeps the original English database.   It introduces the additional challenge of encoding 
cross-lingual correspondences between Chinese and English.\footnote{Please see the appendix for details of the datasets.}
In both datasets, we report  exact set match accuracy, following \citet{yu-etal-2018-spider}. We also report execution accuracy 
in the Spider dataset. 

\subsection{Baselines}

Two kinds of features are widely used in recent semantic parsers 
to boost domain generalization: schema-linking features (as mentioned in Section~\ref{sec:parser})
and pre-trained emebddings such as BERT. To show that our method
can still achieve additional improvements, we compare with strong baselines that are integrated 
with schema-linking features and pre-trained embeddings. 
In the analysis (Section~\ref{subsec:analysis}), we will also show the effect of our method
when both features are absent in the base parsers.

\subsection{Implementation and Hyperparameters}

Our base parser is based on RAT-SQL~\cite{wang2019rat}, which is
implemented in PyTorch~\cite{paszke2019pytorch}.  For English questions and
schemas, we use GloVe~\cite{jeffreypennington2014glove} and BERT-base~\cite{devlin2018bert} 
as the pre-trained embeddings for encoding.  For Chinese questions, we use
Tencent embeddings~\cite{song-etal-2018-directional} and Multilingual-BERT~\cite{devlin2018bert}.
In all experiments, we use a batch size of
$\sSourceDomainBatch = \sTargetDomainBatch = 12 $ and train for up to
20,000 steps.  
See the  Appendix for details on other hyperparameters.

\subsection{Main Results}

Our main results on Spider and Chinese Spider are listed in Table~\ref{table:en-main-results} and 
\ref{table:ch-main-results}, respectively.

\paragraph{Non-BERT Models}
DG-MAML boosts the performance of non-BERT base parsers on Spider and Chinese Spider by \enspidermargin{} and 
\chspidermargin{} respectively, showing its effectiveness in promoting domain generalization.
In comparison, the performance margin for DG-MAML is more significant in
the cross-lingual setting of Chinese Spider. This is presumably due to the fact that heuristic schema-linking 
features, which help promote domain generalization for Spider, are not applicable in Chinese Spider.
We will present more analysis on this in Section~\ref{subsec:analysis}.



\paragraph{BERT Models}
Most importantly, improvements on both datasets are not cancelled out when the base parsers are augmented
with pre-trained representations.
On Spider, the improvements brought by DG-MAML remain roughly the same when the base parser is integrated with BERT-base.
As a result, our base parser augmented with BERT-base and DG-MAML achieves the best execution accuracy compared with previous models.
On Chinese Spider, DG-MAML helps the base parser with multilingual BERT achieve a substantial improvement.
Overall, DG-MAML consistently boosts the performance of the base parser, and is 
complementary to using pre-trained representations.



\begin{table}[t]
    \centering
    \begin{adjustbox}{max width=\columnwidth, center}
        \begin{tabular}{lcc}
            \toprule
            \bfseries Model & \bfseries Dev & \bfseries Test  \\
            \midrule
            \textit{Set Match Accuracy} \\
            SyntaxSQLNet \citep{yu-etal-2018-syntaxsqlnet} & 18.9 & 19.7 \\
            Global-GNN \citep{bogin-etal-2019-global} & 52.7 & 47.4 \\
            IRNet \citep{guo-etal-2019-towards} & 55.4 & 48.5 \\
            RAT-SQL \citep{wang2019rat} & \textbf{62.7} & 57.2 \\
            \textbf{Our Models} \\
            \quad    Base Parser & 56.4 & - \\ 
            \quad    Base Parser + DG-MAML & 58.5 & - \\ 
            \hdashline
            \textit{With BERT-base:} \\
            SyntaxSQLNet + BERT-base  \citep{guo-etal-2019-towards} & 25.0 & 25.4 \\
            IRNet + BERT-base \citep{guo-etal-2019-towards} & 61.9 & 54.7 \\
            BRIDGE + BERT-base \citep{lin-etal-2020-bridging} & 65.5 & 58.2 \\
            RAT-SQL + BERT-base  & 66.0\rlap{$^\diamondsuit$} & - \\
            \textbf{Our Models} \\
            \quad  Base Parser + BERT-base & 66.8 & 63.3 \\ 
            \quad  Base Parser + BERT-base + DG-MAML & \bf 68.9 & \bf 65.2 \\ 
            \hdashline
            \textit{With BERT-large:} \\
            RYANSQL + BERT-large \citep{choi2020ryansql} & \bf 70.6  & 60.6 \\
            RAT-SQL + BERT-large \citep{wang2019rat} & 69.7 & \bf 65.6 \\
            \bottomrule
            \textit{Execution Accuracy} \\
            GAZP + Distil-BERT \citep{zhong-etal-2020-grounded} & 59.2 & 53.5 \\
            BRIDGE + BERT-base \citep{lin-etal-2020-bridging} & 65.3 & 59.9 \\
            \textbf{Our Models} \\
            \quad  Base Parser + BERT-base & 66.8  & 64.1 \\ 
            \quad  Base Parser + BERT-base + DG-MAML & \bf 69.3 & \bf 66.1 \\ 
            \bottomrule
        \end{tabular}
    \end{adjustbox}
    \caption{Accuracy (\%) on the development and test sets of Spider. The first half shows set match accuracy for both 
    non-BERT and BERT models; the second half shows execution accuracy of BERT models.
    Due to the number of model submissions constraint enforced by the Spider team, we only evaluate our BERT models on the test set for now.
    The number with $^\diamondsuit$ is produced by running the code of \citet{wang2019rat}.}
    \label{table:en-main-results}
    \vspace{-5mm}
\end{table}

\subsection{In-Domain vs. Out-of-Domain} 

To confirm that the base parser struggles when applied out-of-domain, we construct
an in-domain setting and measure the gap in performance. This setting also helps us 
address a natural question: does using DG-MAML hurt in-domain performance? This would 
not have been surprising as the parser is explicitly optimized towards better 
performance on unseen target domains. 

To answer these questions, we create a new split of Spider. 
Specifically, for each database from the training and development set of Spider, 
we include 80\% of its question-SQL pairs in the new training set and assign the remaining 20\%  
to the new test set. As a result, the new split consists of 7702 training examples and 1991 test 
examples. When using this split, the parser is tested on databases that all have been seen during training.
We evaluate the non-BERT parsers with the same metric of set match for evaluation.

\paragraph{Does the parser struggle out-of-domain?}
As in-domain and out-of-domain setting have different splits, and thus do not use the same test set,
the direct comparison between them only serves as a proxy to illustrate the effect of domain shift.
We show that, despite the original split of out-of-domain setting containing a larger number of 
training examples (8659 vs 7702), the base parser tested in-domain achieves a much better 
performance (78.2\%) than its counterpart tested out-of-domain (56.4\%). This suggests that the domain shift 
genuinely hurts the base parser. 

\paragraph{Does DG-MAML hurt in-domain performance?}
We study DG-MAML in the in-domain setting to see if it 
hurts in-domain performance.
Somewhat surprisingly, we instead observe a modest improvement (+1.1\%) over the base parser. This suggests that DG-MAML, despite optimizing the model towards domain 
generalization, captures, to a certain degree, a more general notion of generalization or robustness, 
which appears beneficial even in the in-domain setting.

\begin{table}[t!]
    \centering
    \begin{adjustbox}{max width=\columnwidth, center}
        \begin{tabular}{lcc}
            \toprule
            \bfseries Model & \bfseries Dev & \bfseries Test \\
            \midrule
            SyntaxSQLNet \citep{yu-etal-2018-syntaxsqlnet} & 16.4 & 13.3 \\
            \textbf{Our Models} \\
            \quad    Base Parser & 31.0 & 23.0 \\ 
            \quad    Base Parser + DG-MAML & \bf 35.5 & \bf 26.8 \\ 
            \midrule
            \textit{With Multilingual BERT (M-BERT):} \\
            RAT-SQL + M-BERT (Anonymous) & 41.4 & 37.3 \\
            RYANSQL + M-BERT \citep{choi2020ryansql} & 41.3 & 34.7 \\
            \textbf{Our Models} \\
            \quad  Base Parser + M-BERT & 47.0 & 44.3 \\ 
            \quad  Base Parser + M-BERT + DG-MAML & \bf 50.1 & \bf 46.9 \\ 
            \bottomrule
        \end{tabular}
    \end{adjustbox}
    \caption{Set match accuracy (\%) on the development and test sets of Chinese Spider.}
    \label{table:ch-main-results}
    \vspace{-5mm}
\end{table}

\subsection{Additional Experiments and Analysis}
\label{subsec:analysis}

We first discuss additional experiments on linking features and DG-FMAML, 
and then present further analysis probing how DG-MAML works.
As the test sets for both datasets are not publicly available, we will use the
development sets.

\paragraph{Linking Features}

As mentioned in Section \ref{ssec:zsp_compare}, previous work addressed domain generalization
by focusing  on the sub-task of schema linking. For Spider, where questions and schemas are both in English,
\citet{wang2019rat} leverage n-gram matching features which improve  schema linking and significantly 
boost parsing performance.  However, in Chinese Spider, it is not easy and obvious how to design such linking
heuristics. Moreover, as pointed out by \citet{suhr-etal-2020-exploring}, the assumption that columns/tables 
are explicitly mentioned is not general enough, implying that exploiting matching features would not be a good general solution
to domain generalization. Hence, we would like to see \textit{whether DG-MAML can be beneficial when those features are not present}.

Specifically, we consider a variant of the base parser that does not use this feature, 
and train it with conventional supervised learning and with DG-MAML for Spider.  
As shown\footnote{Some results in Table~\ref{table:dg_maml_analyses} differ from Table~\ref{table:en-main-results}. 
The former reports dev set performance over three runs, while the latter shows the best model,
selected based on dev set performance.} in Table~\ref{table:dg_maml_analyses},
we confirm that those features have a big impact on the base parser. More importantly, in the
absence of those features, DG-MAML boosts the performance of the base parser by a larger margin. 
This is consistent with the observation that DG-MAML is more beneficial for Chinese Spider than Spider, in
the sense that the parser would need to rely more on DG-MAML when these 
heuristics are not integrated or not available for domain generalization.

\begin{table}[t!]
    \centering
    \begin{adjustbox}{max width=0.8\columnwidth, center}
        \begin{tabular}{lr}
            \toprule
            \bfseries Model & \bfseries Dev (\%) \\
            \midrule
            \textit{Spider} \\
            \hdashline
            \textbf{Base Parser} & 55.6 $\pm$ 0.5 \\ 
            \quad + DG-FMAML & 56.8 $\pm$ 1.2 \\ 
            \quad  + DG-MAML & \textbf{58.0 $\pm$ 0.8} \\ 
            \hdashline
            \textbf{Base Parser without Features} & 38.2 $\pm$ 1.0 \\ 
            \quad + DG-FMAML & 41.8 $\pm$ 1.5 \\ 
            \quad + DG-MAML & \textbf{43.5 $\pm$ 0.9} \\ 
            \midrule
            \textit{Chinese Spider} \\
            \hdashline
            \textbf{Base Parser} & 29.7 $\pm$ 1.1 \\ 
            \quad + DG-FMAML & 32.5 $\pm$ 1.3 \\ 
            \quad + DG-MAML & \textbf{34.3 $\pm$ 0.9}  \\ 
            \bottomrule
        \end{tabular}
    \end{adjustbox}
    \caption{Accuracy (and $\pm 95\%$ confidence interval) on the development sets of Spider and Chinese Spider.}
    \label{table:dg_maml_analyses}
    \vspace{-5mm}
\end{table}

\paragraph{Effect of DG-FMAML}

We investigate the effect of the first-order approximation in DG-FMAML to see if it would provide 
a reasonable performance compared with DG-MAML. We evaluate it on the development sets of the two datasets, 
see Table~\ref{table:dg_maml_analyses}. DG-FMAML consistently boosts the performance of the base parser, 
although it lags behind \mbox{DG-MAML}. For a fair comparison,
we use the same batch size for DG-MAML and DG-FMAML. 
However, because DG-FMAML uses less memory, it could potentially 
benefit from a  larger batch size.
In practice, DG-FMAML is \textit{twice faster} to train than DG-MAML,
see Appendix for details.

\paragraph{Probing Domain Generalization}

\begin{table}
    \centering
    \small
    \begin{tabular}{lrrr}
        \toprule
        \bfseries Model & \bfseries Precision & \bfseries Recall  & \bfseries F1 \\
        \midrule
        \textit{Spider} \\
        Base Parser & 70.0 & 70.4 &  70.2 \\ 
        Base Parser + DG-MAML & 73.8 & 70.6 & \textbf{72.1} \\ 
        \midrule
        \textit{Chinese Spider} \\
        Base Parser & 61.5 & 60.4 &  61.0 \\ 
        Base Parser + DG-MAML & 66.8 & 61.2 & \textbf{63.9}\\ 
        \bottomrule
    \end{tabular}
    \caption{Performance (\%) of column prediction on the development sets of Spider and Chinese Spider.}
    \label{table:sl_results}
    \vspace{-5mm}
\end{table}
Schema linking has been the focus of previous work on zero-shot semantic parsing.
We take the opposite direction and use this task to probe the parser to see if it, 
at least to a certain degree, achieves domain generalization due to improving schema linking. 
We hypothesize that \textit{improving linking is the mechanism which 
prevents the parser from being trapped in overfitting the source domains}.

We propose to use `relevant column recognition' as a probing task.
Specifically, relevant columns refer to the columns that are mentioned in SQL queries.
For example, the SQL query ``\textit{Select Status, avg(Population) From City Groupby Status}'' 
in Figure \ref{fig:example} contains two relevant columns: `Status' and `Population'.
We formalize this task as a binary classification problem. 
Given a NL question and a column from the corresponding database, 
a  classifier should  predict whether the column is mentioned in the gold SQL query. 
We hypothesize that representations from the DG-MAML parser will be more predictive of relevance than those of the baseline, 
and the probing classifier will detect this difference in the quality of the representations. 

We first obtain the representations for NL questions and schemas from the parsers and keep them fixed.
The binary classifier is then trained based only on these representations.
For classifier training we use the same split as the Spider dataset, i.e.,~the classifier is evaluated on unseen databases.
Details of the classifier are provided in the Appendix.
The results are shown in Table~\ref{table:sl_results}.
The classifier trained on the parser with DG-MAML achieves better performance.
This confirms our hypothesis that using DG-MAML makes the parser have better encodings of NL questions 
and database schemas and that this is one of the mechanisms the parsing model uses to ensure generalization.

\section{Conclusions}
The task of zero-shot semantic parsing has been gaining momentum in recent years.
However, previous work has not proposed algorithms or objectives that explicitly promote domain generalization. 
We rely on the meta-learning framework
to encourage domain generalization.
Instead of learning from individual data points, DG-MAML learns 
from a set of virtual zero-shot parsing tasks.
By optimizing towards better target-domain performance in each simulated task,
DG-MAML encourages the parser to generalize better to unseen domains.

We conduct experiments on two zero-shot text-to-SQL parsing datasets. In both cases,
using DG-MAML leads to a substantial 
boost in performance. 
Furthermore, we show that the faster first-order approximation DG-FMAML can also help a parser achieve better domain generalization.


\section*{Acknowledgements}
We thank Bo Pang, Tao Yu, Qingkai Min and Yuefeng Shi for their help with the evaluation. 
We would like to thank the anonymous reviewers for their valuable comments, and 
Jackie Cheung for pointing out a typo in Eq~\ref{eq:maml_detailed_analysis} in the draft version. We 
gratefully acknowledge the support of the European Research Council (Titov: ERC StG BroadSem 678254; 
Lapata: ERC CoG TransModal 681760) and the Dutch National Science Foundation (NWO VIDI 639.022.518).

\bibliography{anthology,rebibed_main,text2sql-datasets}
\bibliographystyle{acl_natbib}

\appendix
\section{Analysis of DG-FMAML}

Similarly, we use the first-order Taylor expansion to 
analyze the gradients of DG-FMAML:

\begin{align*}
    & \nabla_{\param} \loss_{\tau}(\param) = 
    \nabla_{\param} \loss_{\sSourceDomainBatch}(\param) 
    + \nabla_{\param'} \loss_{\sTargetDomainBatch}(\param') \\
    =& \nabla_{\param} \loss_{\sSourceDomainBatch}(\param) +  
    \nabla_{\param'} \loss_{\sTargetDomainBatch}(\param -  \alpha \nabla_{\param} \loss_{\sSourceDomainBatch}(\param) ) \\
    \approx & \nabla_{\param} \loss_{\sSourceDomainBatch}(\param) + \nabla_{\param'} \loss_{\sTargetDomainBatch}(\param) + \\
    & \quad \alpha \nabla_{\param'}^2 \loss_{\sTargetDomainBatch}(\param) \nabla_{\param} \loss_{\sSourceDomainBatch}(\param) \\
    = & \nabla_{\param} \loss_{\sSourceDomainBatch}(\param) + \nabla_{\param'} \loss_{\sTargetDomainBatch}(\param) + \\
    & \quad \alpha \nabla_{\param'} 
        \big ( 
        \nabla_{\param'} \loss_{\sTargetDomainBatch}(\param) \cdot \nabla_{\param} \loss_{\sSourceDomainBatch}(\param)
        \big )
\label{eq:fmaml_analysis}
\end{align*}

where in Step 3 we expand the gradient function $\nabla_{\param'} \loss_{\sTargetDomainBatch}$ at $\param$. In DG-FMAML, there is no gradients back-propogating from $\param'$ to $\param$, so we can treat $\nabla_{\param} \loss_{\sSourceDomainBatch}(\param)$ and 
$\nabla_{\param'} \loss_{\sTargetDomainBatch}(\param')$ as two independent functions with $\param$ and $\param'$ denoting their parameters respectively. 

In Step 4, the first two terms $\nabla_{\param} \loss_{\sSourceDomainBatch}(\param) + \nabla_{\param'} \loss_{\sTargetDomainBatch}(\param)$ can be viewed as the gradient of applying $\param$ to both source and target domains. 
The last term can be viewed as maximize the dot product between gradients of source and target domain with respect to $\param'$.
In the same spirit as DG-MAML, DG-FMAML also tries to encourage the gradients to generalize between source and target domains.

\section{Datasets}

\paragraph{Spider}

Spider consists of 10,181 examples
(questions and SQL pairs) from 206 databases, including 1,659 examples
taken from the Restaurants
\citep{data-restaurants-original,data-restaurants-logic}, GeoQuery
\citep{data-geography-original}, Scholar
\citep{data-atis-geography-scholar}, Academic \citep{data-academic},
Yelp and IMDB \citep{data-sql-imdb-yelp} datasets.  We follow their
split and use 8,659 examples (from 146 databases) for training, and 1,034 examples (from 20 databases) as our development
set.  The remaining 2,147 examples from 40 test databases are held out and kept by the authors  for evaluation.

\paragraph{Chinese Spider}
Chinese Spider is a Chinese version of Spider that
translates all  NL questions  from English to Chinese and
keeps the original English database.
It simulates the real-life
scenario where  schemas for most relational databases in 
industry are written in English while NL questions from users could be
in any other language. 
Following \citet{min-etal-2019-pilot}, we use the same split of
train/development/test as the Spider dataset.




\section{Hyperparameters}

\paragraph{Base Parser}
We stack 6 relation-aware self-attention layers for encoding. 
Within them, we set the number of attention heads to be 8 and use dropout rate 0.1.
Word embeddings for English questions, column and table names are shared and 
held fixed except for the 50 most common words in the training set.
Word embeddings for Chinese questions are also fixed, 
except for the 50 most common words in the training set. 
As noted in \citet{wang2019rat}, RAT-SQL went through an extensive hyperparameter sweep for 
non-BERT RAT-SQL, which partially explains why our non-BERT base parser is not as 
good as it in Spider. However, after the integration of BERT representations, our base parser 
slightly outperforms RAT-SQL, as shown in the main paper.

\paragraph{Preprocessing}

A major difference between our base parser and RAT-SQL~\cite{wang2019rat}
is the way of preprocessing. During preprocessing, input questions, column names and table names in
schemas are tokenized and lemmatized by Stanza~\cite{qi2020stanza}
which can handle both English and Chinese.

\paragraph{Learning Rates}
We use the learning rate of $\alpha = 5 \times 10^{-4}$
for meta-train.  For the final update of parameters, we use
Adam~\citep{kingma2014adam} with the learning rate $6 \times 10^{-4}$. 
We manually search for the best meta-train learning rates from 
$1 \times 10^{-4}$ to $9 \times 10^{-4}$ with the step size  $2 \times 10^{-4}$,
based on performance on the development set.
Other hyperparameters are not tuned.
For the learning rate of final update (not $\alpha$ of meta-train), we use the same scheduler as \citet{wang2019rat}.
Specifically, during the first 500 steps, the learning rate linearly increases from 0 to $6 \times 10^{-4}$.  
Then, it is annealed to 0 with $6 \times 10^{-4} (1 - \frac{step - 500}{9500})^{-0.5}$.


\paragraph{Hardware and Model Size}
Our non-BERT models are trained using NVIDIA GeForce RTX 2080, which has a memory size of 11GB. 
The base parser has around 10 million parameters, 
where around 1.5 million parameters are pre-trained embeddings that are mostly fixed during training.
For BERT models, we first find the best hyperparameters using GeForce RTX 2080 with a small batch size; then 
we train them using V100 to save cost.

\paragraph{Average Runtime}
The average training time for the non-BERT base parser, DG-MAML and DG-FMAML are 10, 24, 13 hours per run.
For BERT models, the numbers are 36, 68, 42 hours per run. 

\subsection{Loss Curve}

\begin{figure*}[t!]
    \centering
    \begin{subfigure}[t]{0.5\textwidth}
        \centering
        \includegraphics[width=0.98\textwidth]{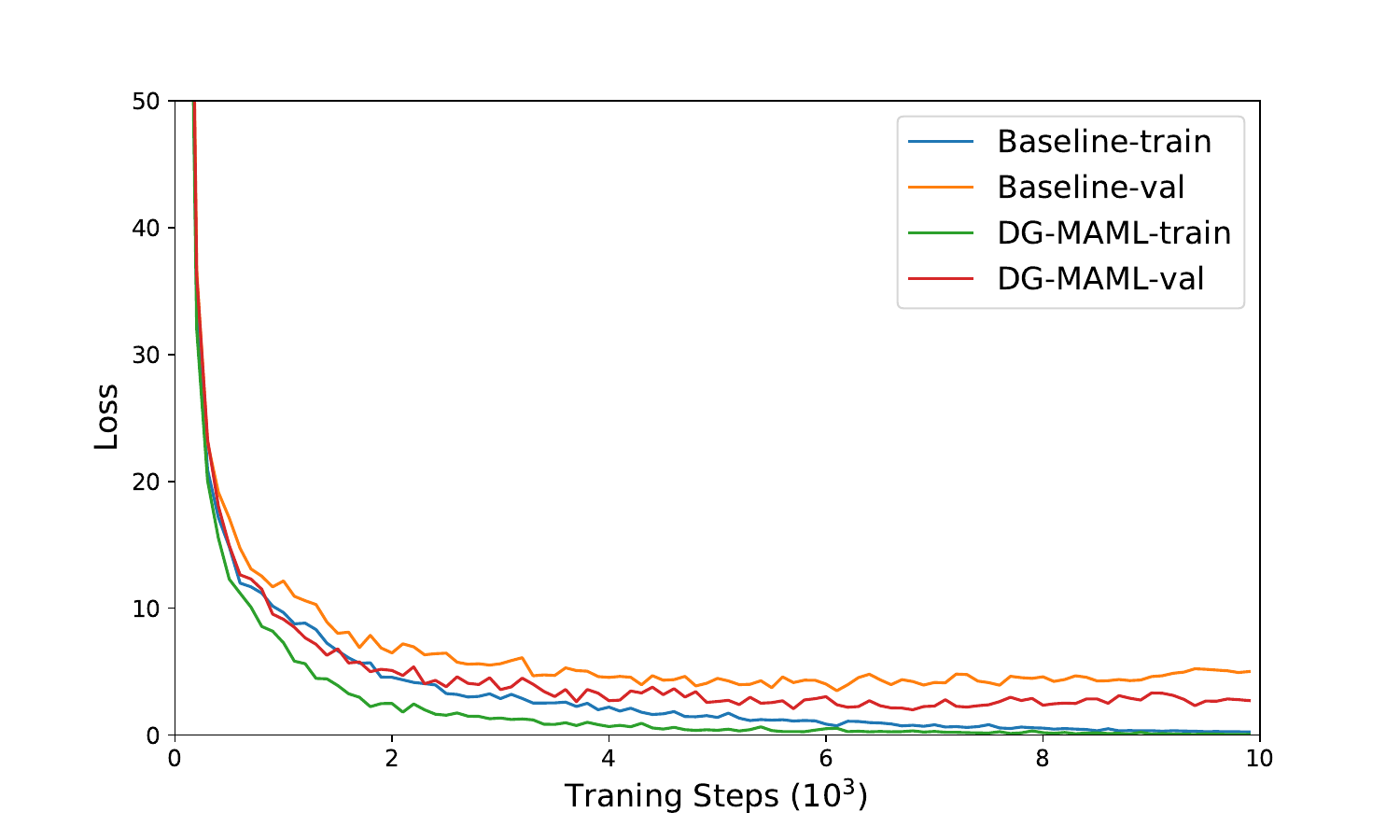}
        \caption{Loss curves on the Spider dataset.}
    \end{subfigure}%
    ~ 
    \begin{subfigure}[t]{0.5\textwidth}
        \centering
        \includegraphics[width=0.98\textwidth]{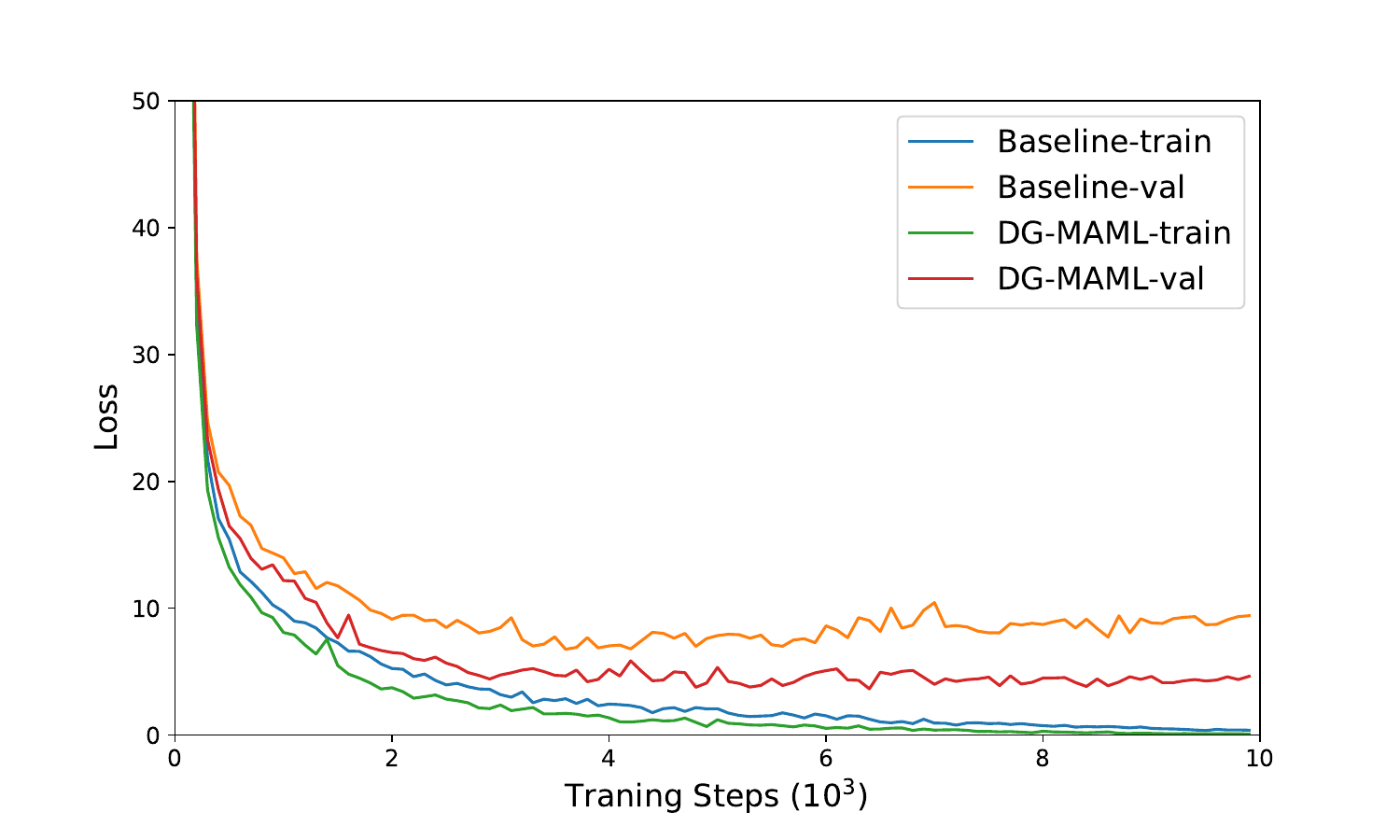}
        \caption{Loss curves on the Chinese Spider dataset.}
    \end{subfigure}
    \caption{Comparison of losses when the parser is trainined with conventaional supervised learning (Baseline) and DG-MAML.}
    \label{fig:loss}
\end{figure*}

In Figure \ref{fig:loss}, we show the loss curves of the models on the two datasets during training.
In comparison, DG-MAML helps to reduce the gap between training and validation loss. 

\section{Classifier for Probing Domain Generalization}

The classifier takes the input of a pair of (column, question) and outputs a binary label indicating whether the column is relevant. 
As explained in the paper, we retrieve the representations of columns and questions from a pre-trained parser.
We denote the representation of a column as $\vc \in \R^{k}$, and the representation of a question as $\vq \in \R^{n \times k}$ 
where $n$ is the number of words in the question and $k$ is the size of encoding.

For each pair of $(\vc, \vq)$, we first align the column $\vc$ softly with the question $\vq$ using an attention function, 
and obtain an aligned representation $\vt$ for the column. 
Then we compute a score of relevance based on the aligned representation.
Finally, a probability $p$ of relevance is computed through a sigmoid function $\sigmoid$.

\begin{align}
    \begin{split}
        \vt &= \attention(\vc, \vq) \\
        score &= \mlp(\vc, \vt) \\
        p &= \sigmoid(score)
    \end{split}
\end{align}

\end{document}